\documentclass[a4paper,12pt]{article}
\usepackage{geometry}
\usepackage{fontspec}
\usepackage{setspace}
\usepackage{fancyhdr}
\usepackage{ctex}
\usepackage[english]{babel}
\usepackage{booktabs}
\usepackage{graphicx}
\usepackage{amsmath, amssymb, mathtools}
\usepackage{algorithm}
\usepackage{float}
\usepackage{algpseudocode}
\usepackage{bm}
\usepackage{times}
\usepackage{url}
\usepackage[sort&compress]{natbib}
\numberwithin{equation}{section}
\setcitestyle{numbers,square,comma}
\usepackage{hyperref}
\usepackage{multirow}

\newcommand{\tbf}{\textbf} 
\newcommand{\sums}{\sum\limits}
\newcommand{\lims}{\lim\limits}
\newcommand{\dd}{\ \mathrm{d}}
\newcommand{\res}{\mathrm{res}}

\newcommand{\R}{\mathbb{R}}
\newcommand{\PP}{\mathbb{P}}

\newcommand{\ci}{\mathcal{C}}
\newcommand{\li}{\mathcal{L}}
\newcommand{\ji}{\mathcal{J}}

\newcommand{\f}{\bm{f}}
\newcommand{\x}{\bm{x}}
\newcommand{\z}{\bm{z}}
\newcommand{\X}{\bm{X}}
\newcommand{\Z}{\bm{Z}}
\newcommand{\bta}{\bm{\theta}}

\usepackage{cancel}

\begin{document}
\title{Solving stationary Fokker-Planck equations via adaptive sampling from solution-induced and residual-guided distributions \thanks{This work was supported by the National Natural Science Foundation of China (under grant 12401566).}}
\author{
Jiangfan Chen \thanks{Email: 240320055@fzu.edu.cn}\quad
Li Zeng\thanks{Corresponding author.
Email: lzeng@fzu.edu.cn}\\[0.5em]
{\small School of Mathematics and Statistics, Fuzhou University,}
{\small Fuzhou, 350108, China}
}
\date{}
\maketitle
\begin{abstract}
	We propose Solution-Residual Adaptive Sampling (SRAS) for solving stationary Fokker-Planck equations (SFPEs) on unbounded domains using KRnet. Normalizing flows ensure normalization and non-negativity by construction, but the governing equation must still be enforced through training. The selection of training points can therefore affect the accuracy of the resulting solution. SRAS combines samples from the current density approximation with samples from Gaussian mixtures centered at points with large partial differential equation (PDE) residuals, balancing probability-mass coverage with residual-guided refinement. Numerical experiments on five problems ranging from two to eight dimensions show that SRAS generally reduces the errors in fewer adaptive iterations and improves the density approximation accuracy compared with adaptive deep density approximation and residual-based adaptive refinement. 
	
\noindent\tbf{Keywords} Stationary Fokker-Planck Equation; Normalizing Flow; Adaptive Deep Density Approximation; Solution-Residual Adaptive Sampling
\par\medskip
\end{abstract}

\section{Introduction}\label{Introduction}
	The stationary Fokker-Planck equation (SFPE) characterizes stationary probability densities of diffusion processes and plays an important role in the study of stochastic dynamical systems. Its numerical solution presents several challenges: the solution must be non-negative and integrate to one, the spatial domain is often unbounded, and the state space can be high-dimensional. Classical grid-based methods, such as finite difference and finite element methods, require careful discretization to preserve these probabilistic constraints. On unbounded domains, they typically also require domain truncation and suitable artificial boundary conditions, while their computational cost grows rapidly with dimension. 

	In recent years, deep learning has shown considerable potential for solving partial differential equations (PDEs). Representative methods include physics-informed neural networks (PINNs) \cite{PINN} and the deep Galerkin method \cite{DeepGalerkin}, which minimize losses based on differential-equation residuals and associated conditions; the deep Ritz method \cite{DeepRitz}, which minimizes variational energy functionals; and weak adversarial networks \cite{WAN}, which formulate weak solutions through minimax optimization. Within this broader landscape, PINNs have become an established approach to physics-informed machine learning \cite{PINNScene1}. Physics-informed methods have also been developed for forward and inverse stochastic problems \cite{PINNSolveSDE1, PINNSolveSDE2}. For SFPEs, existing approaches include direct neural approximation of stationary densities \cite{PINNSolveFPE} and score-based methods for high-dimensional problems \cite{Score-FPE}. In the former approach, normalization is imposed through a penalty term in the training loss \cite{PINNSolveFPE}, rather than guaranteed by the density representation itself.

	The solution of an SFPE is a probability density function and must therefore be non-negative and integrate to one. These constraints motivate the use of normalizing flows (NFs), which represent normalized, non-negative densities by construction \cite{NF1, NF2}. Through an invertible, sufficiently regular transformation of a prescribed base distribution, NFs allow explicit density evaluation via the change-of-variables formula and direct sampling \cite{NF2}. These properties have motivated NF-based solvers for FPEs \cite{NFSolveFPE1, ANF-FPE, FlowKac}. Among these methods, adaptive deep density approximation (ADDA) uses KRnet, a normalizing flow based on invertible block-triangular mappings \cite{KRnet}, to approximate stationary densities \cite{ADDA-KR}. We also adopt KRnet as the density representation in this work. Although normalization and non-negativity are satisfied by construction, the governing SFPE must still be enforced through training. In collocation-based residual minimization, the sampling distribution influences optimization and, unless importance weights preserve a prescribed integration measure, the residual objective being approximated \cite{ImportanceSamplingPINN, RAD}. Adaptive sampling has therefore been investigated to improve training-point selection in both NF-based density approximation and PINN-based PDE solvers \cite{ADDA-KR, DAS-PINN, RAD}.
	
	For FPEs, the probabilistic structure of the solution provides a natural basis for adaptive training-point selection. In ADDA, collocation points are drawn from the current KRnet density approximation, concentrating training effort in regions of high predicted probability mass \cite{ADDA-KR}. Related adaptive density-approximation strategies have been developed for time-dependent FPEs using temporal NFs \cite{tKRnet} and for fractional FPEs using simplified KRnet models \cite{FFPEAdaptive}. For bounded domains, bounded KRnet combines invertible flows on hypercubes with adaptive sampling to approximate PDE solutions that are probability density functions \cite{BoundedKRnet}.
	
	On the other hand, residual-based adaptive sampling has been studied extensively in the PINN literature for training-point selection \cite{RAR, RAD, R3Sampling}. In collocation-based residual formulations,  both PINNs and NF-based PDE solvers are trained by minimizing PDE-residual losses at sampled points, so sampling principles developed for PINNs can also inform NF-based PDE solvers. Residual-based sampling strategies direct samples toward regions with large PDE residuals. For instance, residual-based adaptive refinement (RAR) periodically adds candidate points with the largest residual magnitudes to the training set \cite{RAR}. Residual-based adaptive distribution  and residual-based adaptive refinement with distribution  construct sampling distributions from functions of the PDE residual \cite{RAD}. The retain-resample-release  strategy retains points whose residuals exceed the mean residual of the current sample and resamples the remaining points, while allowing retained points to be released when they no longer satisfy this criterion \cite{R3Sampling}. DAS-PINNs approximate a residual-induced distribution using KRnet \cite{DAS-PINN}, whereas Gaussian mixture distribution-based adaptive sampling  constructs a Gaussian mixture distribution from residual information to guide sampling \cite{GAS-PINN}. Failure-informed PINNs  instead identify residual-defined failure regions to guide sample placement \cite{FI-PINN1, FI-PINN2}.

	Solution-induced and residual-based sampling emphasize different aspects of the current approximation when selecting training points. Sampling from the current density approximation emphasizes regions of high predicted probability mass but may undersample regions where the density is underestimated. Residual-based sampling targets local violations of the governing equation, yet a large pointwise residual does not necessarily indicate a large contribution to the overall density error. These considerations motivate Solution-Residual Adaptive Sampling (SRAS), combining solution-induced sampling with residual-guided exploration. Building on residual-guided Gaussian mixture sampling \cite{GAS-PINN}, SRAS uses the current density approximation both to generate training points and to define an adaptive candidate region for residual exploration. Points with large residuals in this region serve as Gaussian mixture centers, providing additional samples for local refinement. A subset of existing training points and samples from the initial distribution are also included to help stabilize training. We evaluate SRAS against ADDA and RAR on five stationary problems in two to eight dimensions, focusing on density accuracy and error reduction over adaptive iterations.

	The rest of the paper is organized as follows. Section \ref{Problem setup} formulates the stationary Fokker-Planck problem. Section \ref{Methodology} introduces the normalizing-flow framework and develops the SRAS strategy. Section \ref{Numerical study} presents numerical experiments comparing SRAS with ADDA and RAR. Section \ref{Conclusions} concludes the paper. Appendix \ref{Appendix A} outlines the KRnet architecture.

\section{Problem setup}\label{Problem setup}
	In this section, we describe the problem under consideration. Assume that $(\Omega, \mathcal{F}, \{\mathcal{F}_t\}_{t\geq0}, \mathbb{P})$ is a complete filtered probability space satisfying the usual hypotheses. Let $\bm{W}_t: \Omega\to \R^d$ be a standard Wiener process adapted to this filtration. Consider the stochastic process $\X_t: \Omega\to\R^d$ governed by the It\^{o} stochastic differential equation
\begin{equation}
	\dd \bm{X}_t = \bm{\mu}(\bm{X}_t, t)\dd t + \bm{\sigma}(\bm{X}_t, t) \dd \bm{W}_t,
\end{equation}
where the drift vector $\bm{\mu}: \R^d \times [0, \infty)\to\R^d$ and the diffusion matrix $\bm{\sigma}: \R^d \times [0, \infty)\to\R^{d\times d}$ are Borel measurable. Assume that for each $t\in [0, \infty)$, the probability distribution of $\X_t$ is absolutely continuous with respect to Lebesgue measure. By the Radon-Nikodym theorem, this distribution admits a probability density function $p(\cdot, t): \R^d\to \R$, defined up to equality almost everywhere. Formally, $p(\x, t)$ satisfies the Fokker-Planck equation (FPE) \cite{FPEWellPosedness}
\begin{equation}\label{FPE}
	\dfrac{\partial p(\bm{x}, t)}{\partial t} =  \li[p](\bm{x}, t) := -\nabla\cdot \big(\bm{\mu}(\bm{x}, t) p(\bm{x}, t) \big) + \nabla\cdot\big(\nabla\cdot\big( \bm{D}(\bm{x}, t) p(\bm{x}, t) \big) \big),
\end{equation}
with the normalization and non-negativity conditions
\begin{equation}
	\int_{\R^d} p(\x,t)  \dd \x = 1, \quad p(\x, t)\geq0,
\end{equation}
together with the initial condition and the decay condition at infinity
\begin{equation}\label{IBC}
	p(\x, 0) = p_0(\x), \quad \lims_{|\x|\to\infty}p(\x, t) = 0,
\end{equation}
where $| \cdot |$ denotes the Euclidean norm, and the diffusion tensor $\bm{D}: \R^d\times [0, \infty) \to \R^{d\times d}$ is given by
$$
	D_{ij}(\x, t)={\dfrac{1}{2}}\sums_{k=1}^{d}\sigma _{i k}(\x, t)\sigma _{j k}(\x, t).
$$
	
	In this work, we consider the time-homogeneous case, in which $\bm{\mu}$ and $\bm{\sigma}$ depend only on $\x$, and seek a stationary probability density. We assume that the associated SFPE admits a unique classical probability density solution $p(\x)$ satisfying the decay condition at infinity. As a time-independent solution of the FPE, $p(\x)$ satisfies
\begin{equation}\label{SFPEs}
	\begin{dcases}
	\li[p](\bm{x}) =  -\nabla\cdot\big( \bm{\mu}(\bm{x}) p(\bm{x}) \big) + \nabla\cdot\big(\nabla\cdot \big( \bm{D}(\bm{x}) p(\bm{x}) \big) \big) = 0, \\
	\int_{\R^d} p(\bm{x}) \dd \x = 1,\quad p(\bm{x})\geq 0, \\
	\lims_{|\x|\to\infty} p(\bm{x}) = 0.
	\end{dcases}
\end{equation}

\section{Methodology}\label{Methodology}
	In this section, we present a normalizing-flow-based framework for approximating the stationary probability density satisfying SFPE \eqref{SFPEs}. We use KRnet to represent the approximate density, ensuring normalization and non-negativity by construction, and train its parameters by minimizing the strong-form PDE residual at sampled collocation points. This mesh-free formulation avoids an explicit spatial grid, but its accuracy depends in part on the placement of training points, particularly on unbounded domains and in higher dimensions. To guide their placement, we develop Solution-Residual Adaptive Sampling (SRAS), which combines sampling from the current density approximation with residual-guided Gaussian mixture sampling. The resulting strategy allocates training points to regions of high predicted probability mass and to neighborhoods of selected points with large PDE residuals.

\subsection{Normalizing Flow Framework}	
\label{sec:NF_framework}
	A normalizing flow defines a tractable probability density through an invertible transformation of a prescribed base distribution \cite{NF1, NF2}. In this work, we use this representation to approximate the stationary probability density satisfying SFPE \eqref{SFPEs}.

	Let \(\mathbb{P}_{\X}\) denote the target distribution with unknown continuous density \(p_{\X}\), and let \(\mathbb{P}_{\Z}\) denote a prescribed base distribution with known continuous density \(p_{\Z}\), such as a Gaussian density. Suppose that there exists a \(C^1\)-diffeomorphism \(\f^*:\mathbb R^d\to\mathbb R^d\) that pushes forward $\mathbb P_{\Z}$ to $\mathbb P_{\X}$, namely, $(\f^*)_\# \mathbb{P}_{\Z} = \mathbb{P}_{\X}$. Here, the pushforward measure $(\f^*)_\# \mathbb{P}_{\Z}$ is defined by
	
\begin{equation*}
	\big( (\f^*)_\# \mathbb{P}_{\Z} \big)(B)\coloneqq \mathbb{P}_{\Z}\big( (\f^*)^{-1}(B) \big), \quad B\in \mathcal{B}(\R^d),
\end{equation*}
where \(\mathcal B(\mathbb R^d)\) denotes the Borel \(\sigma\)-algebra on \(\mathbb{R}^d\). 
By the change-of-variables theorem for densities, the unknown density $p_{\X}$ can be expressed as
\begin{equation}\label{change of variable}
	p_{\X}(\x) = p_{\Z}\big((\f^*)^{-1}(\x)\big) \big| \det \nabla (\f^*)^{-1}(\x) \big|, \quad \x\in\R^d.
\end{equation}

Since \(\f^*\) is unknown, an NF uses a trainable \(C^1\)-diffeomorphism \(\f_{\bm{\theta}}\), parameterized by \(\bm{\theta}\), to approximate the target distribution. To obtain a flexible representation while retaining tractable density evaluation and sampling, \(\f_{\bm{\theta}}\) is usually constructed as a composition of invertible layers: 
\begin{equation*}
\f_{\bm{\theta}}=\f_{L}\circ\cdots\circ\f_1,
\end{equation*}
where each \(\f_l:\mathbb R^d\to\mathbb R^d\) is a \(C^1\)-diffeomorphism, and \(\bm{\theta}\) collects the trainable parameters of the layers. Each \(\f_l\) is designed for efficient evaluation of forward and inverse maps and their Jacobian determinants, making density evaluation and sample generation tractable for the full composition. The inverse mapping is given by
\begin{equation*}
\f_{\bm{\theta}}^{-1}=\f_1^{-1}\circ\cdots\circ \f_L^{-1}.
\end{equation*}

For \(\Z\sim \mathbb P_Z\), the model random vector
\(\X_{\bm{\theta}}=\f_{\bm{\theta}}(\Z)\) has distribution \(\mathbb P_{\bm{\theta}}=(\f_{\bm{\theta}})_\#\mathbb P_{\Z}\) and density
\begin{equation}
p_{\bm{\theta}}(\x)=p_{\Z}\bigl(\f_{\bm{\theta}}^{-1}(\x)\bigr)
\left|\det \nabla \f_{\bm{\theta}}^{-1}(\x)\right|=p_{\Z}\bigl(\f_{\bm{\theta}}^{-1}(\x)\bigr)
\left|\det \nabla \f_{\bm{\theta}}(\f_{\bm{\theta}}^{-1}(\x))\right|^{-1}.
\end{equation}
This density is normalized and non-negative by construction and serves as an approximation to \(p_X\). For a given input \(\z\in\mathbb R^d\), define the intermediate variables
\[\z_0=\z,\qquad \z_l=\f_l(\z_{l-1}),\qquad l=1,\ldots,L.\]
The chain rule then yields
\[
\left|\det \nabla \f_{\bm{\theta}}(\z)\right|
=
\prod_{l=1}^{L}
\left|\det \nabla \f_l(\z_{l-1})\right|.
\]
To generate samples from \(P_\theta\), we draw independent samples from the base distribution and apply the forward mapping:
\begin{equation}
\z^{i}\overset{\mathrm{i.i.d.}}{\sim} \mathbb P_{\Z},\qquad
\x^{i}=\f_{\bm{\theta}}\bigl(\z^{i}\bigr),
\qquad i=1,\ldots,N.
\end{equation}
The resulting samples follow the model distribution \(\mathbb P_{\bm{\theta}}\), which approximates \(\mathbb P_{\X}\).

A \(C^1\)-diffeomorphism is sufficient to define the model density through the change-of-variables formula. To evaluate the strong-form SFPE residual, we further require \(p_{\bm{\theta}}\in C^2(\mathbb R^d)\). A sufficient condition is that the base density \(p_{\Z}\) belongs to \(C^2(\mathbb R^d)\) and \(\f_{\bm{\theta}}\) is a \(C^3\)-diffeomorphism. 
In this work, we adopt KRnet \cite{KRnet}, which extends the RealNVP architecture \cite{realNVP}. To satisfy the smoothness requirement for strong-form residual evaluation, we use smooth activation functions in the coupling networks and omit the piecewise-defined nonlinear layer. The remaining layers are used with parameterizations that preserve invertibility.  A brief outline of the KRnet architecture is provided in Appendix \ref{Appendix A}, while further details can be found in the original work \cite{KRnet}.
 
\subsection{Adaptive Sampling Strategy}

	Using the density representation introduced in Section \ref{sec:NF_framework}, we determine the model parameters by minimizing a sampled strong-form SFPE residual. 
	Let \(\Theta\) denote the admissible parameter space. For a fixed probability measure \(\mathbb P\) on \(\mathbb{R}^d\), define the loss function:
	\begin{equation}
		\label{eqn:continuous_PINN_loss}
	\mathcal J(\bm{\theta};\mathbb{P})=\mathbb E_{\x\sim \mathbb{P}}\!\left[r_{\bm{\theta}}(\x)^2\right]=\int_{\mathbb R^d}r_{\bm{\theta}}(\x)^2\,\mathrm{d}\mathbb P,
	\end{equation}
	where \(r_{\bm{\theta}}(\x)=\left|\mathcal{L}[p_{\bm{\theta}}](\x)\right|\) is the pointwise SFPE residual.

	We assume $\ji(\bm{\theta};\mathbb{P})$ is finite and approximate it using Monte Carlo sampling. Given \(N\) independent samples $\ci=\{\x^{i}\}_{i=1}^{N}$ from $\mathbb{P}$, the empirical loss is
	\begin{equation}
		\label{eqn:empirical_loss}
		\widehat{\ji}(\bm{\theta}; \ci) =\dfrac{1}{|\ci|} \sums_{\x\in \ci} r_{\bm{\theta}}^2(\x),
	\end{equation}
	During training, we alternate between optimizing the model on a fixed collocation set and updating that set using the trained model. The choice of the sampling distribution is therefore an important part of the training procedure.

	Sampling from the current model distribution provides a natural starting point. This strategy allocates training points according to the predicted probability mass and has been used in previous NF-based PDE solvers \cite{ADDA-KR, FFPEAdaptive, tKRnet, BoundedKRnet}. However, regions where the current model underestimates the density may receive too few points. In particular, solution-induced sampling alone may provide insufficient coverage of poorly resolved modes or tail regions. Residual-based sampling offers another criterion by directing points toward locations where the current approximation violates the governing equation. Such strategies have been investigated extensively for PINNs \cite{RAR, RAD, R3Sampling, DAS-PINN, GAS-PINN, FI-PINN1, FI-PINN2}. Nevertheless, the magnitude of a local PDE residual does not directly quantify the corresponding density error. Concentrating samples solely according to residual  may therefore neglect other regions that are important for the overall density approximation. These two criteria emphasize different aspects of the current approximation. We therefore combine solution-induced sampling with residual-guided refinement. To retain information from previous training stages and maintain sampling in regions covered by the initial distribution, we also retain a subset of the existing training points and draw additional points from the initial training distribution.

	Let \(\ci_k=\{\x_k^i\}_{i=1}^{N}\) denote the training set at adaptive stage \(k\), and let \(\bm{\theta}_k\) denote the parameters obtained after training on \(\ci_k\). Define the empirical measure 
	\begin{equation*}
	\widehat{\mathbb P}_{\ci_k}=\frac1N\sum_{i=1}^{N}\delta_{\x_k^i},
	\end{equation*}
	where \(\delta_{\x}\) is the Dirac measure at \(\x\). We construct the next training set using the target mixture 
	\begin{equation}
	\PP_{k+1} = \gamma_0 \PP_0 + \gamma_1\widehat{\mathbb P}_{\ci_k}+\gamma_2\PP_{\bta_k}+\gamma_3\PP_k^\res
	\qquad\gamma_i>0,\qquad\sum_{i=0}^{3}\gamma_i=1.
	\end{equation}
	Here, \(\mathbb P_0\) is the initial training distribution, \(\mathbb P_{\boldsymbol\theta_k}\) is the updated model distribution, and \(\mathbb P^\res_k\) is a residual-based distribution defined below. The weights \(\gamma_i\) control the contributions of the four components. By linearity of integration, the loss function \eqref{eqn:continuous_PINN_loss} under the target mixture \(\PP_{k+1}\) can be expressed as
	\begin{equation*}
	\mathcal J(\bm{\theta};\mathbb{P}_{k+1})=\gamma_0 \mathcal J(\bm{\theta};\PP_0)+\gamma_1\mathcal J(\bm{\theta};\widehat{\mathbb P}_{\ci_k})+\gamma_2\mathcal J(\bm{\theta};\PP_{\bta_k}) + \gamma_3 \mathcal J(\bm{\theta};\PP_k^\res).
	\end{equation*}
	The sampling measure is held fixed during the subsequent optimization stage. In practice, the mixture weights determine the allocation of collocation points among the four components, and training uses the empirical loss evaluated on the combined training set.
	Samples for the first three mixture components are obtained by drawing from \(\PP_0\), selecting points uniformly without replacement from \(\ci_k\), and applying the forward mapping described in Section \ref{sec:NF_framework} to base samples to generate samples from \(\PP_{\bm{\theta}_k}\), respectively. It remains to construct the residual-guided distribution \(\PP_k^{\mathrm{res}}\).
	To this end, we use a residual-guided Gaussian mixture, whose centers are selected within a region determined by the current model. 
	
	First, draw \(N_{\text{c}}\) auxiliary samples $\{\bm{y}_k^i\}_{i=1}^{N_\text{c}}$ from the updated model distribution \(\PP_{\bm{\theta}_k}\).  For each coordinate $j = 1, \dots, d$, let $\widehat Q_{j, k}(\alpha)$ denote the empirical $\alpha$-quantile of $\{(\bm{y}_k^i)_j\}_{i=1}^{N_\text{c}}$. Given a quantile parameter $0<\delta<1 / 2$, define $a_{j, k} = \widehat Q_{j,k}(\delta)$ and $b_{j,k} = \widehat Q_{j,k}(1-\delta)$. We expand each interquantile interval at both endpoints by a fraction $\rho > 0$ of its width and define the search region
$$
	\mathcal{B}_k(\delta,\rho) = \prod_{j=1}^{d}\left[a_{j,k}-\rho\bigl(b_{j,k}-a_{j,k}\bigr), b_{j,k}+\rho\bigl(b_{j,k}-a_{j,k}\bigr)\right].
$$
The box adapts to the coordinatewise spread of the model samples and is intended to capture the bulk of the model probability mass. 

	Next, we map $N_\text{c}$ Sobol quasi-random points \cite{Sobol} into \(\mathcal{B}_k(\delta,\rho) \) to form a candidate set $S_k$. We then evaluate \(r_{\bm{\theta}_k}\) at these candidates and select the $m$ points $\{\x_\res^{i}\}_{i=1}^m$ with the largest residuals, where \(1\leq m\leq N_{\text{c}}\). Define $\PP_{k}^\res$ as the Gaussian mixture distribution:
$$
	\PP_{k}^\res := \dfrac{1}{m}\sums_{i=1}^m \mathcal{N}(\x_\res^{i}, \sigma^2 \bm{I}_d),
$$
where $\sigma>0$ is the Gaussian standard deviation that controls the local exploration scale. A sample from \(\PP_{k}^\res\) is obtained by selecting a center uniformly and adding an independent perturbation drawn from \(\mathcal{N}(0, \sigma^2I_d)\). Thus, residual-guided sampling explores neighborhoods of the selected centers rather than repeatedly using the centers themselves. The generated samples may lie outside \(\mathcal{B}_k(\delta,\rho)\), although the search for centers is restricted to this model-informed region.
	
	We now combine the four sampling components to construct the next training set \(\ci_{k+1}\). For the prescribed training-set size \(N\), set 
	\(n_i=\lceil\gamma_iN\rceil\), \(i=0,1,2,3\). We draw \(n_0\) points from \(\mathbb P_0\), select \(n_1\) points uniformly without replacement from \(\ci_k\), draw \(n_2\) points from \(\mathbb P_{\bm{\theta}_k}\), and draw \(n_3\) points from \(\PP_{k}^\res\). These four collections are pooled, and \(N\) entries are selected uniformly without replacement to form \(\ci_{k+1}\). The resulting training set is used to evaluate 
	\begin{equation}
		\label{eqn:empirical_loss_at_k}
		\widehat{\ji}(\bm{\theta}; \ci_{k+1}) =\dfrac{1}{|\ci_{k+1}|} \sums_{\x\in \ci_{k+1}} r_{\bm{\theta}}^2(\x),
	\end{equation}
    with equal weight assigned to each collocation point. The overall adaptive sampling measure combines solution-induced coverage with residual-guided refinement, and we refer to this strategy as \textbf{Solution-Residual Adaptive Sampling (SRAS)}.

	 Training begins with an initialized KRnet $p_{\bta}$ and \(N\) points drawn from \(\PP_0\). At each adaptive iteration $k$, we perform an optimization stage followed by a refinement stage. In the optimization stage, we minimize the empirical loss $\widehat{\ji}(\bta; \ci_k)$ over the training set $\ci_k$ to update the model parameters $\bta$. In the refinement stage, we use SRAS to generate a new training set $\ci_{k+1}$. The refinement stage is omitted at the final iteration since no further optimization follows. The overall training procedure is summarized in Algorithm \ref{Algorithm 1}.

	\begin{algorithm}[h]
	\caption{Solution-Residual Adaptive Sampling (SRAS) for the SFPE}
	\label{Algorithm 1}
	\textbf{Input:}
	Initialized KRnet $p_{\bta}$, initial training distribution $\PP_0$, number of adaptive iterations $N_\text{a}$, number of training epochs per iteration $N_\text{s}$, number of training points $N$, batch size $m_\text{b}$, learning-rate schedule $\{\eta_k\}_{k=0}^{N_{\mathrm a}-1}$, mixture weights $\{\gamma_i\}_{i=0}^{3}$, quantile parameter $\delta$, box expansion factor $\rho$, number of candidate points $N_\text{c}$, number of exploration points $m$, Gaussian standard deviation $\sigma$.
	\begin{algorithmic}[1]
	\State Draw $N$ points from $\PP_0$ to form $\ci_0$.
	\State Set $n_i=\lceil\gamma_iN\rceil$, $i=0, 1, 2, 3$.
	\For{$k=0,\ldots,N_{\mathrm a}-1$}
		\State Set the optimizer learning rate to $\eta_k$.
		\State Randomly partition the training set $\ci_{k}$ into $n_\text{b}= \lceil N/m_\text{b}\rceil$ mini-batches $\{\ci_{k}^{(i_\text{b})}\}_{i_\text{b} =0}^{n_\text{b}-1}$.
		\For{$j = 0$ to $N_\text{s} - 1$}
			\For{$i_\text{b} = 0$ to $n_\text{b} - 1$}
	             \State Compute the empirical loss $\widehat{\ji}(\bta; \ci_{k}^{(i_\text{b})})$ by Eq.~\eqref{eqn:empirical_loss}.
	             \State Update the parameters $\bta$ using an optimizer with learning rate $\eta_k$.
			\EndFor
	    \EndFor
		\State Denote the updated parameters by $\bm{\theta}_k$.
		\If{$k<N_{\mathrm a}-1$}
			\State Draw $N_{\mathrm c}$ auxiliary samples from
			$\PP_{\bm{\theta}_k}$ and construct $\mathcal{B}_k(\delta,\rho)$.
			\State Generate $N_\text{c}$ Sobol points $S_k$ in $\mathcal{B}_k(\delta,\rho)$
			and select the $m$ points with the largest
			$r_{\bm{\theta}_k}$ to define $\PP_k^{\mathrm{res}}$.
			\State Draw $n_0$ points from $\PP_0$,
			$n_2$ points from $\PP_{\bm{\theta}_k}$,
			and $n_3$ points from $\PP_k^{\mathrm{res}}$.
			\State Select $n_1$ points uniformly without replacement
			from $\ci_k$.
			\State Pool the four collections and select $N$ entries
			uniformly without replacement to form $\ci_{k+1}$.
		\EndIf
	\EndFor
	\end{algorithmic}
	\textbf{Output:} Trained density $p_{\bm{\theta}_{N_{\mathrm a}-1}}$.
	\end{algorithm}

\section{Numerical study}\label{Numerical study}
	This section evaluates the SRAS strategy on five stationary Fokker-Planck problems in dimensions ranging from two to eight. Let $p^*$ denote the exact reference density, $p_{\bm{\theta}}$ the model density, and $\mathbb{P}^*$ the distribution with density $p^*$. We measure density approximation accuracy using the relative $L^2$ error and the relative Kullback--Leibler (KL) divergence, defined as follows:
\begin{align}
	\mathcal{E}_{L^2}(p^*, p_{\bm{\theta}}) = \dfrac{\left\| p^* - p_{\bm{\theta}}\right\|_{L^2(\R^d)}}{\left\| p^*\right\|_{L^2(\R^d)}}, \quad 
	\mathcal{E}_{\text{KL}}(p^*, p_{\bm{\theta}}) = \dfrac{D_{\text{KL}}(p^*\| p_{\bm{\theta}})}{H(p^*)}.
	\label{eqn:measure_errors}
\end{align}
Here, $D_{\mathrm{KL}}(p^*\|p_{\bm{\theta}})$ is the forward KL divergence and $H(p^*)$ is the differential entropy,
\begin{equation}
D_{\text{KL}}(p^*\| p_{\bm{\theta}})=\int_{\mathbb{R}^d}p^*(\x)\log\left(\dfrac{p^*(\bm{x})}{p_{\bm{\theta}}(\bm{x})}\right)\mathrm{d}\bm{x}, \quad H(p^*)=-\int_{\mathbb{R}^d}p^*(\x)\log p^*(\x)\mathrm{d}\x.
\end{equation}

	We approximate these metrics using Monte Carlo sampling. Specifically, we draw an evaluation set $\ci_\text{v}=\{\x^{i}\}_{i=1}^{N_\text{e}}$ independently from $\PP^*$ and approximate the quantities \eqref{eqn:measure_errors} by
\begin{align*}
	\widehat{\mathcal{E}}_{L^2}(p^*, p_{\bm{\theta}}) &= \left( \dfrac{\sum_{i=1}^{N_\text{e}} \left( \left|p^*(\x^{i}) - p_{\bm{\theta}}(\x^{i})\right|^2 / p^*(\x^{i})  \right)}{\sum_{i=1}^{N_\text{e}}p^*(\x^{i})}\right)^{\frac{1}{2}},\\
	\widehat{\mathcal{E}}_{\text{KL}}(p^*, p_{\bm{\theta}}) &= \dfrac{\sum_{i=1}^{N_\text{e}} \left( \log {p^*(\x^{i})} - \log{p_{\bm{\theta}}(\x^{i})}\right)}{-\sum_{i=1}^{N_\text{e}} \log {p^*(\x^{i})}}.
\end{align*}

	We compare SRAS with two baselines: the solution-induced ADDA strategy \cite{ADDA-KR} and the residual-based RAR strategy \cite{RAR}. For each problem, all  three methods use the same network architecture and optimization hyperparameters. They differ in the rule used to update the training set $\ci_{k+1}$ at adaptive iteration $k$. The sampling rules used in this work are as follows:
	\begin{itemize}
    \item \textbf{ADDA}: Draw $N$ points from the solution-induced distribution $\PP_{\bta}$ to form $\ci_{k+1}$.
    \item \textbf{RAR}: Draw $N$ candidate points from the initial training distribution $\PP_0$, select the $\lceil 0.05N \rceil$ points with the largest residuals $r_{\bm{\theta}_k}$, and append them to \(\ci_k\) to form \(\ci_{k+1} \).
    \item \textbf{SRAS}: Construct \(\ci_{k+1}\) using Algorithm~\ref{Algorithm 1} with the mixture parameters specified for each problem.
	\end{itemize}

	ADDA and SRAS maintain \(N\) training points, whereas RAR increases the training-set size after each refinement. Consequently, equal numbers of training epochs do not imply equal numbers of optimizer updates or equal computational costs. We therefore compare the methods in terms of error reduction over adaptive iterations.

	The common experimental settings are as follows. For KRnet, we choose the $d$-dimensional standard Gaussian $\mathcal{N}(0, \bm{I}_d)$ as the base distribution $\PP_{\Z}$. Each affine coupling layer is parameterized by a feed-forward network with hyperbolic tangent (tanh) activations. For training, we draw the initial training points from $\PP_0 = \mathcal{U}([-5, 5]^d)$, initialize all trainable parameters using Glorot initialization \cite{Glorot}, and optimize the model using the SOAP optimizer \cite{SOAP}. For evaluation, we use $N_\text{e} = 100{,}000$ evaluation points. All numerical experiments are implemented in PyTorch.

\subsection{Two-dimensional bimodal distribution I}
\begin{figure}[h]
	\centering
	\includegraphics[width=0.35\textwidth]{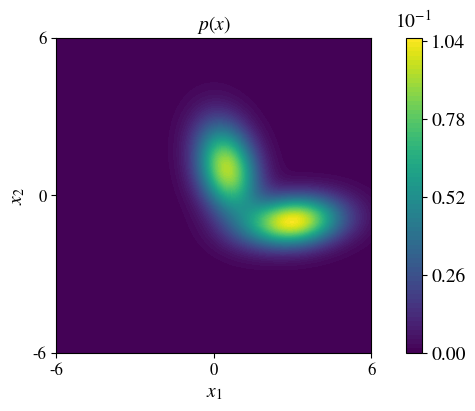}
	\caption{Exact solution.}
	\label{2d bimodal I exact solution}
\end{figure}

We first consider the two-dimensional SFPE \eqref{SFPEs} with the drift and diffusion specified below.
	
	\begin{equation}\label{2d bimodal I}
		\bm{\mu}(\x) = \nabla \log \hat{p}(\x)  + 2 \bm{v}\big(\bm{v}^\top \nabla \hat{p}(\x)\big),\quad
		\bm{D}(\x) = \bm{I}_2 + \bm{v}\bm{v}^\top \hat{p}(\x), 
	\end{equation}
	where 
\begin{equation}
\begin{split}
	\hat{p}(\x) = \beta_1\mathcal{N}(\x; \bm{\mu}_1, \bm{\Sigma}_1) + \beta_2 &\mathcal{N}(\x; \bm{\mu}_2, \bm{\Sigma}_2),\ \bm{v} = (3.913, 0.846), \\
	\beta_1 = \beta_2 = 0.5, \quad &\bm{\mu}_1 = (3, -1),\ \bm{\mu}_2 = (0.5, 1),\\
     \bm{\Sigma}_1 = 
    \begin{bmatrix}
    1.5 & 0.1 \\
    0.1 & 0.4
    \end{bmatrix}, \quad 
    &\bm{\Sigma}_2 = 
    \begin{bmatrix}
    0.6 & -0.2 \\
    -0.2 & 1.3
    \end{bmatrix}.
\end{split}
\end{equation}
The exact solution is $p^*(\x) = \beta_1\mathcal{N}(\x; \bm{\mu}_1, \bm{\Sigma}_1) + \beta_2 \mathcal{N}(\x; \bm{\mu}_2, \bm{\Sigma}_2)$, as shown in Figure \ref{2d bimodal I exact solution}.

	We approximate the solution by a KRnet with \(8\) affine coupling layers, each parameterized by a network with two hidden layers of \(48\) neurons each. The initial training set contains \(N=20{,}000\) points, and the mini-batch size is \(m_b=1{,}000\). Training comprises $N_\text{a} = 50$ adaptive iterations and $N_\text{s} = 100$ epochs per iteration. The learning rate is $\eta = 0.0001$ in the first adaptive iteration and 0.0005 thereafter. For SRAS, we set the mixture weights to $\gamma_0 = 0.05$, $\gamma_1 = 0.1$, $\gamma_2  = 0.5$, and $\gamma_3 = 0.35$. The quantile parameter is $\delta = 0.1$, and the box expansion factor is $\rho = 0.1$. We generate $N_\text{c} = 20{,}000$ candidate points and select $m = 2{,}000$ with the largest residuals as Gaussian mixture centers.

\begin{table}[h]
  \centering
  \caption{Final relative $L^2$ errors and relative KL divergences for different sampling strategies.}
  \label{2d bimodal I final errors}
  \begin{tabular}{@{}lccccccc@{}}
    \toprule
    & \multirow{2}{*}{ADDA} & \multirow{2}{*}{RAR} & \multicolumn{5}{c}{SRAS\ ($\sigma$)} \\
    \cmidrule(lr){4-8}
    & & & 0.01 & 0.05 & 0.1 & 0.5 & 1 \\
    \midrule
    $\widehat{\mathcal{E}}_{L^2}$ & 8.502e-02 & 5.021e-02 & 2.720e-02 & 3.372e-02 & \tbf{2.613e-02} & 2.910e-02 & 3.403e-02 \\
    $\widehat{\mathcal{E}}_{\mathrm{KL}}$ & 2.874e-03 & 8.105e-04 & 3.840e-04 & 6.004e-04 & \tbf{2.930e-04} & 3.969e-04 & 4.516e-04 \\
    \bottomrule
  \end{tabular}
\end{table}
	
\begin{figure}[H]
	\centering
	\includegraphics[width=0.7\textwidth]{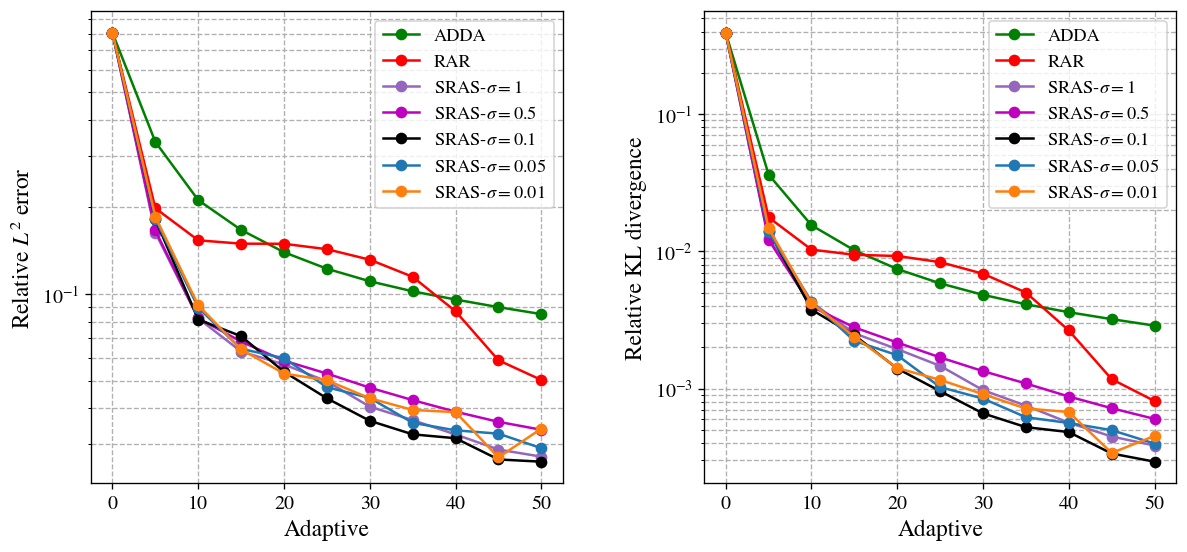}
	\caption{Evolution of the relative $L^2$ error and relative KL divergence for different sampling strategies.}
	\label{2d bimodal I errors}
\end{figure}

	Table \ref{2d bimodal I final errors} reports the final relative $L^2$ errors and relative KL divergences obtained with ADDA, RAR, and SRAS for $\sigma \in \{1, 0.5, 0.1, 0.05, 0.01\}$. All tested SRAS configurations yield smaller errors than ADDA and RAR. Among the SRAS variants, $\sigma=0.1$ gives the smallest relative $L^2$ error and relative KL divergence. The dependence on \(\sigma\) is not monotone, and reducing \(\sigma\) below \(0.1\) does not improve the final accuracy.	Figure \ref{2d bimodal I errors} shows the evolution of the relative $L^2$ error and relative KL divergence over the adaptive iterations. ADDA reduces the errors gradually. RAR improves more slowly during the intermediate stages but exhibits a larger reduction near the end of training, finishing below ADDA but not the SRAS variants. Moreover, RAR appends newly selected residual points to the existing training set at each update, so the number of training points, and hence the training work per epoch, increases progressively. SRAS reaches lower error levels earlier and maintains an advantage over both baselines.
	 
\begin{figure}[H]
	\centering
	\includegraphics[width=0.8\textwidth]{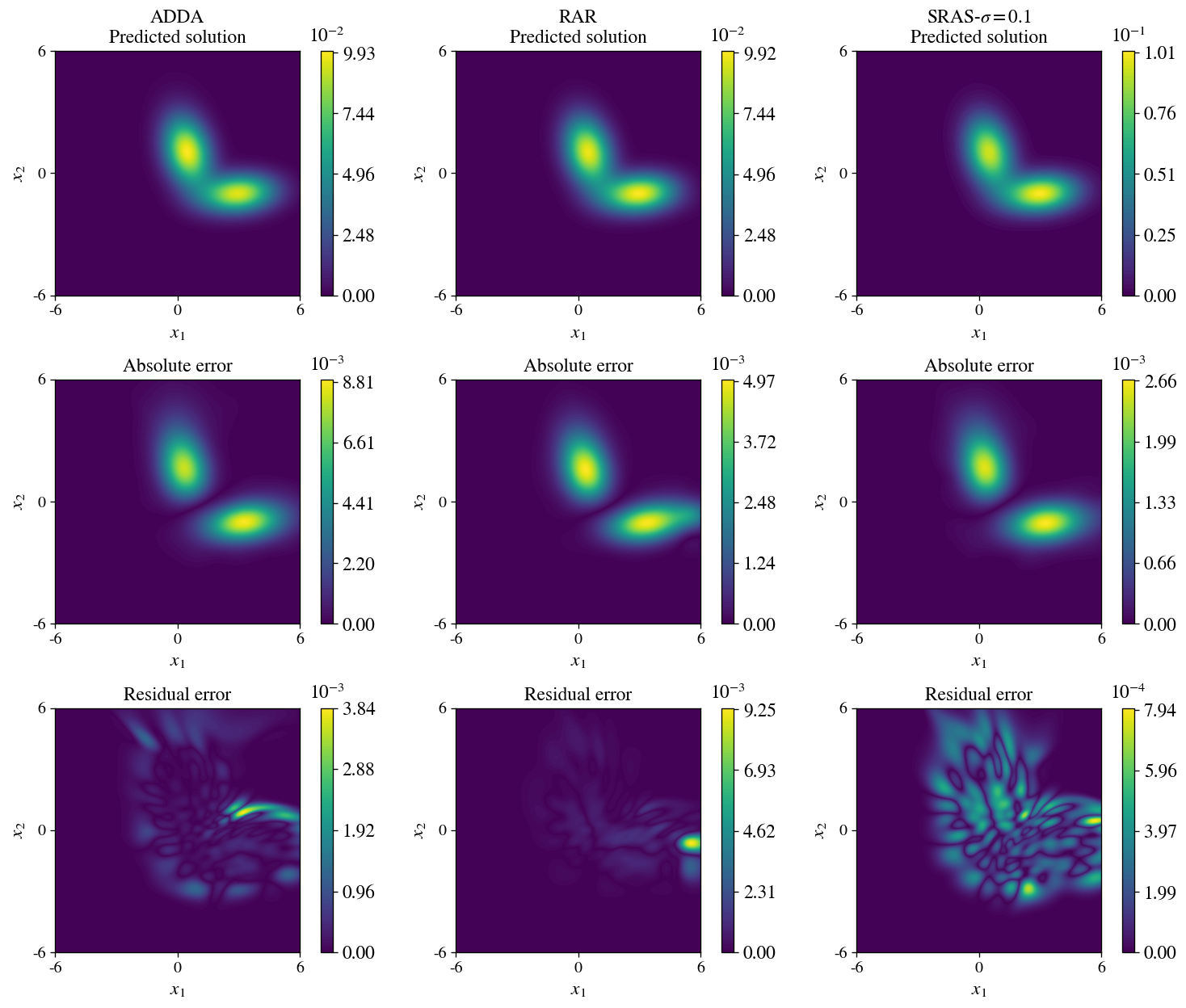}
	\caption{Predicted solutions, pointwise absolute errors, and pointwise residuals for different sampling strategies.}
	\label{2d bimodal I predict}
\end{figure}

	Figure \ref{2d bimodal I predict} compares the predicted solutions, pointwise absolute errors, and pointwise residuals for ADDA, RAR, and SRAS with $\sigma=0.1$. All three methods recover the bimodal structure. On the displayed domain, SRAS exhibits smaller maximum absolute errors and residuals, consistent with its lower reported relative $L^2$ error and relative KL divergence. The residual pattern for SRAS also appears more spatially dispersed, with less pronounced localization than those of ADDA and RAR.

\begin{figure}[H]
	\centering
	\includegraphics[width=0.80\textwidth]{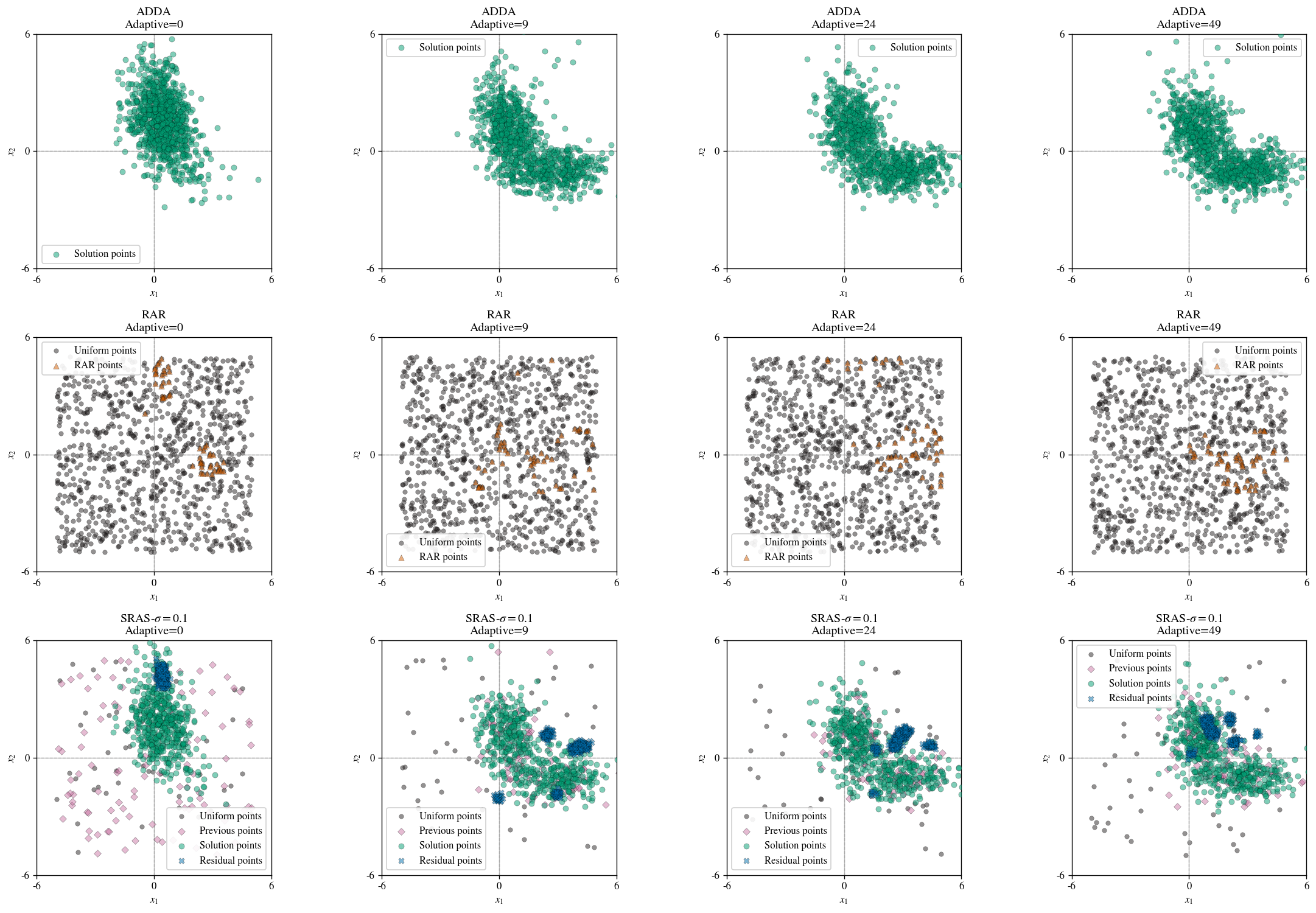}
	\caption{Samples for ADDA, RAR, and SRAS ($\sigma=0.1$).}
	\label{2d bimodal I samples1}
\end{figure}

	Figure \ref{2d bimodal I samples1} presents samples for ADDA, RAR, and SRAS with $\sigma=0.1$ at the four displayed adaptive iterations labeled $0$, $9$, $24$, and $49$. For visualization, \(1{,}000\) points are shown for each method at each displayed iteration.
	For SRAS, these points are allocated among the four sampling components in proportion to their prescribed weights. 
	The ADDA samples initially concentrate near one modal region and subsequently cover both modal regions.
	For RAR, the residual-selected points occupy different locations within the uniform sampling background. For SRAS, the solution-induced samples cover the predicted high-probability regions, while the residual-guided samples form localized clusters whose locations and spatial extent change across iterations.
	These changes illustrate how the sampling patterns adapt as training proceeds.
	
\begin{figure}[H]
	\centering
	\includegraphics[width=0.80\textwidth]{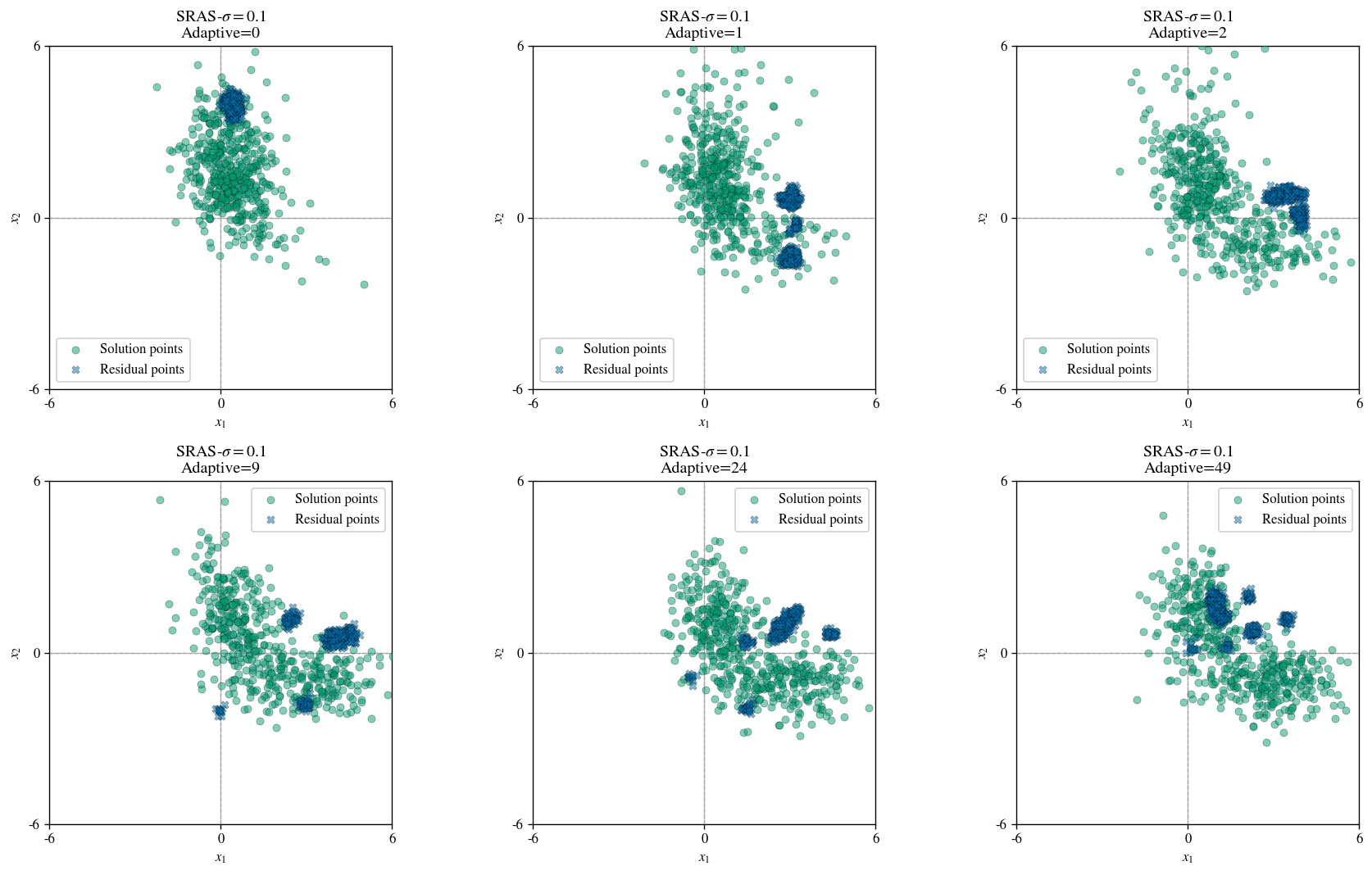}
	\caption{Solution-induced and residual-guided sampling components of SRAS with $\sigma=0.1$.}
	\label{2d bimodal I samples2}
\end{figure}

	Figure \ref{2d bimodal I samples2} displays only the solution-induced and residual-guided components of SRAS with \(\sigma=0.1\). 	Each panel displays $500$ solution-induced points and $350$ residual-guided points, corresponding to their respective shares of the 1,000-point visualization sample. The solution-induced points cover regions of predicted probability mass, whereas the residual-guided points form localized clusters around selected high-residual centers. Some of these clusters lie near the modal regions and in the region between them, providing additional collocation points to constrain the PDE residual in the transition region.

\subsection{Two-dimensional bimodal distribution II}
We next consider a two-dimensional Gaussian mixture with unequal component weights and a more spatially dispersed density than in the first example. The SFPE has constant isotropic diffusion,
with coefficients given by
\begin{equation}\label{2d bimodal II}
	\begin{split}
	\bm{\mu}(\x) = \nabla \log\big(\beta_3\mathcal{N}(\x; \bm{\mu}_3, \bm{\Sigma}_3) + \beta_4\mathcal{N}(\x; \bm{\mu}_4, \bm{\Sigma}_4)\big),\quad
	\bm{D}(\x) = \bm{I}_2,
	\end{split}
\end{equation}
where 
\begin{equation}\label{2d bimodal II parameter}
	\begin{split}
    \beta_3 = 0.55, \ \beta_4 = 0.45, \quad &\bm{\mu}_3 = (-1, -1),\ \bm{\mu}_4 = (3, 3),\\
     \bm{\Sigma}_3 = 
    \begin{bmatrix}
    6.1219 & -0.2637 \\
    -0.2637 & 1.8166
    \end{bmatrix}, \quad 
    &\bm{\Sigma}_4 = 
    \begin{bmatrix}
    5.7657 & -1.4047 \\
    -1.4047 & 5.3840
    \end{bmatrix}.
	\end{split}
\end{equation}
The exact solution is $p^*(\x) = \beta_3\mathcal{N}(\x; \bm{\mu}_3, \bm{\Sigma}_3) + \beta_4\mathcal{N}(\x; \bm{\mu}_4, \bm{\Sigma}_4)$.

	We use the same network architecture, initial training-set size, mini-batch size, number of adaptive iterations, epochs per iteration, and learning-rate schedule as in the first example. For SRAS, we set the mixture weights to $\gamma_0=0.05$, $\gamma_1=0.1$, $\gamma_2=0.65$, and $\gamma_3=0.2$. The quantile parameter and box expansion factor remain $\delta=\rho=0.1$. We generate $N_{\mathrm c}=20{,}000$ candidate points and select the $m=1{,}000$ points with the largest residuals as Gaussian mixture centers. We test $\sigma\in\{0.01, 0.05, 0.1, 0.5, 1\}$.

\begin{figure}[h]
	\centering
	\includegraphics[width=0.70\textwidth]{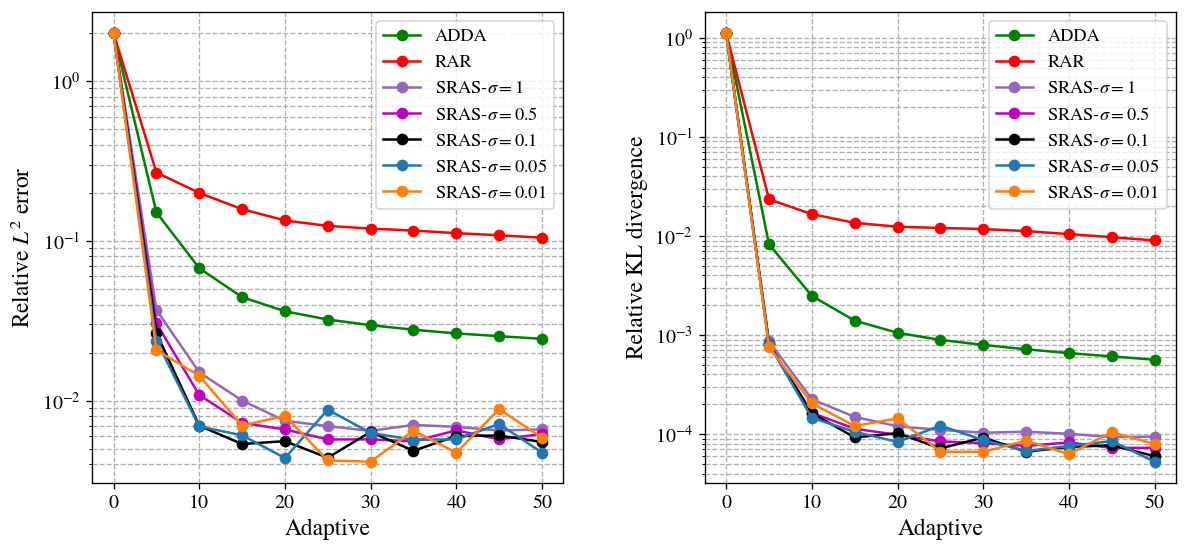}
	\caption{Evolution of the relative $L^2$ error and relative KL divergence for different sampling strategies.}
	\label{2d bimodal II errors}
\end{figure}

	Figure \ref{2d bimodal II errors} shows the evolution of the relative $L^2$ error and relative KL divergence over the adaptive iterations. All methods exhibit a rapid initial decrease, but their subsequent behavior differs. ADDA gradually reduces both metrics and maintains lower errors than RAR during the later iterations. RAR also continues to improve but retains the largest late-stage errors. The SRAS variants reach substantially lower error levels within fewer adaptive iterations. Their errors subsequently fluctuate around low levels, with limited further decrease, particularly in the relative \(L^2\) error for smaller \(\sigma\). Table \ref{2d bimodal II final errors} reports the errors averaged over the last ten adaptive iterations for all methods. Unlike the first example, ADDA outperforms RAR in both reported metrics, while all tested SRAS variants yield lower mean errors than both baselines. Among the tested values, $\sigma=0.05$ gives the smallest mean errors in both metrics, although the differences among several
SRAS configurations are small.
	
\begin{table}[h]
  \centering
  \caption{Late-stage mean relative $L^2$ errors and relative KL divergences over the last ten adaptive iterations for different sampling strategies.}
  \label{2d bimodal II final errors}
  \begin{tabular}{@{}lccccccc@{}}
    \toprule
    & \multirow{2}{*}{ADDA} & \multirow{2}{*}{RAR} & \multicolumn{5}{c}{SRAS\ ($\sigma$)} \\
    \cmidrule(lr){4-8}
    & & & 0.01 & 0.05 & 0.1 & 0.5 & 1 \\
    \midrule
    $\widehat{\mathcal{E}}_{L^2}$ & 2.523e-02 & 1.084e-01 & 6.093e-03  & \tbf{5.977e-03} & 6.175e-03 & 6.068e-03 & 6.735e-03\\
    $\widehat{\mathcal{E}}_{\mathrm{KL}}$ & 6.057e-04 & 9.724e-03 & 7.473e-05 & \tbf{7.219e-05} & 7.541e-05 & 7.468e-05 & 9.625e-05 \\
    \bottomrule
  \end{tabular}
\end{table}

\subsection{Three-dimensional non-Gaussian distribution}
\begin{figure}[h]
	\centering
	\includegraphics[width=0.80\textwidth]{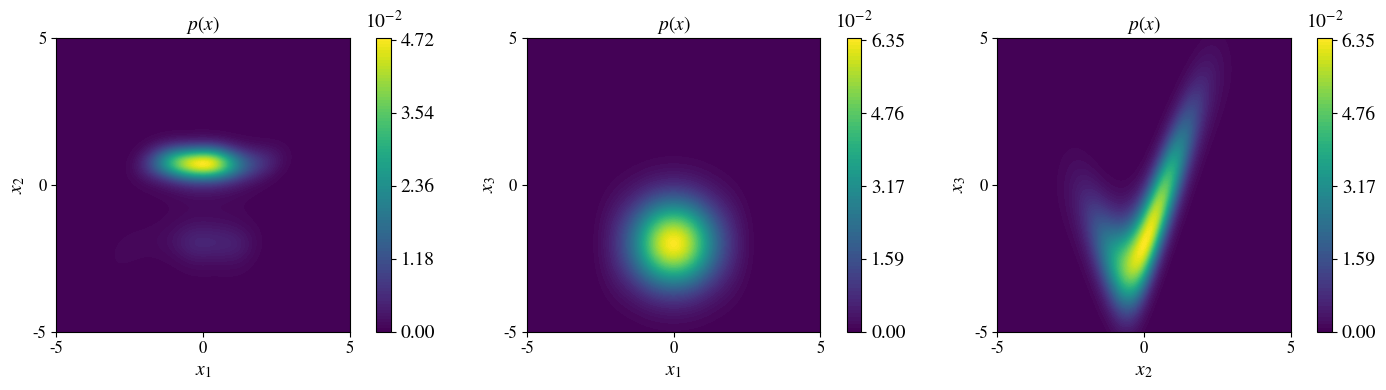}
	\caption{Exact solution: two-dimensional views of the three-dimensional density.}
	\label{3d single exact solution}
\end{figure}

We next consider a three-dimensional non-Gaussian stationary density with nonlinear dependence among its coordinates. The drift and diffusion in the SFPE \eqref{SFPEs} are given by
\begin{equation}\label{3d bimodal}
\begin{split}
	\bm{\mu}(\x) = 2\x - \bm{D}(\x)\nabla H(\x),\quad
	\bm{D}(\x) = \bm{I}_3 + \mathrm{diag}\{x_1^2, x_2^2, x_3^2\}, 
\end{split}
\end{equation}
where
\begin{equation}
\begin{split}
	H(\x) = \dfrac{x_1^2 + ( x_2 - 2\tanh(x_1) + 2\sin(x_1) )^2 + (x_3 - 2\tanh(x_2) + 2\cos(x_2) )^2}{2}.
\end{split}
\end{equation}
The exact solution is $p^*(\x) = (2\pi)^{-\frac{3}{2}}\exp\{-H(\x)\}$, which has a unique mode at \((0,0,-2)\). Figure \ref{3d single exact solution} presents its two-dimensional views.

	We set \(K=3\) and use $L=8$ affine coupling layers per outer stage, each parameterized by a network with two hidden layers of \(72\) neurons each. 
We start with $N = 45{,}000$ training points with a batch size of $m_\text{b} = 1{,}500$. Training comprises $N_\text{a} = 20$ adaptive iterations with $N_\text{s} = 100$ epochs per iteration. The learning rate is fixed at $\eta = 0.0001$ throughout training. For SRAS, we set the mixture weights to $\gamma_0 = 0.05$, $\gamma_1 = 0.1$, $\gamma_2  = 0.6$, and $\gamma_3 = 0.25$. The quantile parameter and box expansion factor are \(\delta=\rho=0.05\). We generate $N_\text{c} = 30{,}000$ candidate points and select $m = 3{,}000$ with the largest residuals as Gaussian mixture centers. The Gaussian standard deviation is \(\sigma=0.1\).
	
\begin{figure}[h]
	\centering
	\includegraphics[width=0.70\textwidth]{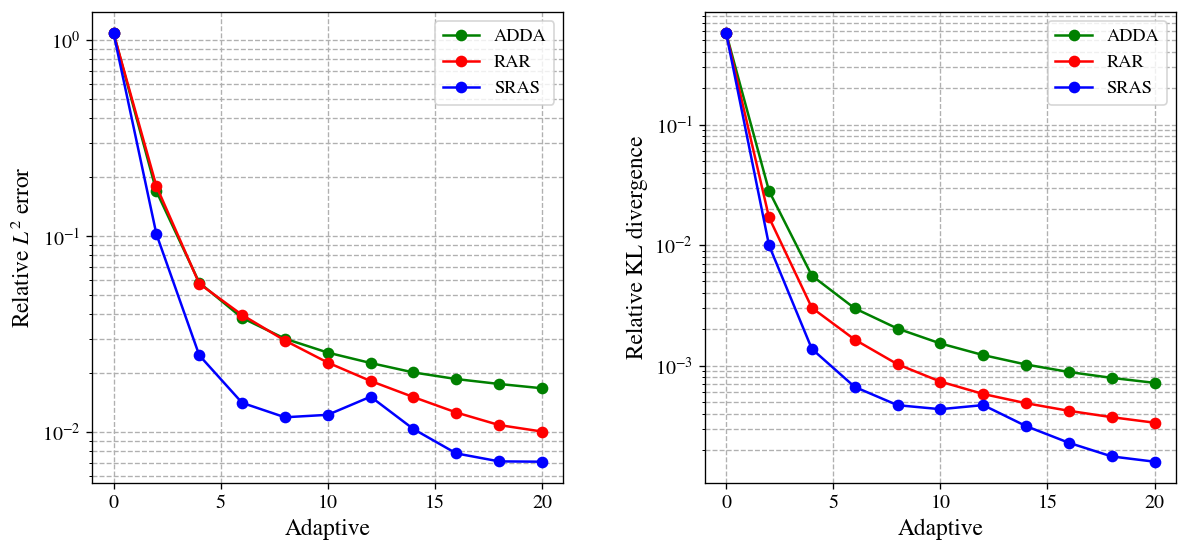}
	\caption{Evolution of the relative $L^2$ error and relative KL divergence for different sampling strategies.}
	\label{3d single errors}
\end{figure}
	
	Figure \ref{3d single errors} shows the evolution of the relative $L^2$ error and relative KL divergence over the adaptive iterations. All three methods reduce both metrics substantially during the initial adaptive updates. ADDA subsequently improves more slowly, while RAR continues to reduce the errors and attains lower final values than ADDA. SRAS generally maintains the lowest error levels and achieves the smallest final errors in both metrics, despite a temporal increase in its relative \(L^2\) error during the intermediate iterations. These results demonstrate the effectiveness of SRAS for this non-Gaussian problem with state-dependent diffusion.

\subsection{Four-dimensional bimodal distribution}
We next consider a four-dimensional SFPE with state-dependent diffusion. The drift and diffusion in the SFPE~\eqref{SFPEs} are given by
\begin{equation}\label{4d bimodal}
\begin{split}
	\bm{\mu}(\x) &= 2\x + \bm{D}(\x)\nabla \log\big(\beta_5\mathcal{N}(\x; \bm{\mu}_5, \bm{\Sigma}_5) + \beta_6\mathcal{N}(\x; \bm{\mu}_6, \bm{\Sigma}_6)\big),\\
	\bm{D}(\x) &= \bm{I}_4 + \mathrm{diag}\{x_1^2, x_2^2, x_3^2, x_4^2\}, 
\end{split}
\end{equation}
where
\begin{equation}
\begin{split}
	\beta_5 = \beta_6 = 0.5, \quad \bm{\mu}_5 = (-1, -1.5, -0.5, 0),\ \bm{\mu}_6 = (0.5, 0.5, 0.5, 0.5),\\
	\bm{\Sigma}_5 = 
	\begin{bmatrix}
	2.5 & 0.1 & 0 & 0\\
      0.1 & 1 & -0.2 & 0 \\
      0 & -0.2 & 2.3 & 0 \\
	0 & 0 & 0 & 0.5
	\end{bmatrix}, \quad 
	\bm{\Sigma}_6 = 
	\begin{bmatrix}
	0.5 & 0 & 0 & 0 \\
	0 & 1.8 & -0.1 & 0.5 \\
      0 & -0.1 & 2 & 0.25 \\
      0 & 0.5 & 0.25 & 0.9
	\end{bmatrix}.
\end{split}
\end{equation}
The exact solution is $p^*(\x) = \beta_5\mathcal{N}(\x; \bm{\mu}_5, \bm{\Sigma}_5) + \beta_6\mathcal{N}(\x; \bm{\mu}_6, \bm{\Sigma}_6)$.

	We set $K=4$ and use $L=8$ affine coupling layers per outer stage, each parameterized by a network with two hidden layers of \(96\) neurons each. We start with $N = 80{,}000$ training points with a batch size of $m_\text{b} = 2{,}000$. Training comprises $N_\text{a} = 10$ adaptive iterations and $N_\text{s} = 50$ epochs per iteration. The learning rate is $\eta = 0.0001$ in the first adaptive iteration and 0.0005 thereafter. For SRAS, we set the mixture weights to $\gamma_0 = 0.05$, $\gamma_1 = 0.1$, $\gamma_2  = 0.6$, and $\gamma_3 = 0.25$. The quantile parameter and the box expansion factor are $\delta = \rho = 0.05$. We generate $N_\text{c} = 40{,}000$ candidate points and select $m = 2{,}000$ with the largest residuals as Gaussian mixture centers. The Gaussian standard deviation is $\sigma = 0.1$.
	
\begin{figure}[h]
	\centering
	\includegraphics[width=0.70\textwidth]{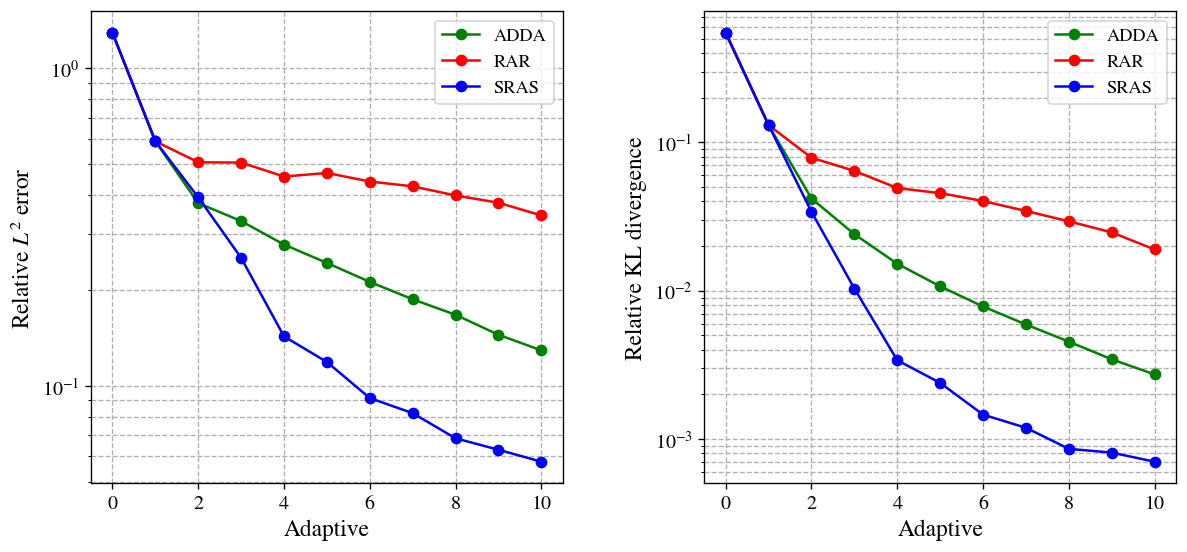}
	\caption{Evolution of the relative $L^2$ error and relative KL divergence for different sampling strategies.}
	\label{4d bimodal errors}
\end{figure}

	Figure \ref{4d bimodal errors} shows the evolution of the relative $L^2$ error and relative KL divergence over the adaptive iterations. ADDA reduces both metrics steadily, whereas RAR shows the slowest decrease and remains at the highest error levels. After the initial iterations, SRAS reduces both metrics more rapidly than the two baselines and attains the lowest final errors. 
\subsection{Eight-dimensional bimodal distribution}
	Finally, we consider an eight-dimensional bimodal stationary density with constant isotropic diffusion. The drift and diffusion in the SFPE~\eqref{SFPEs} are given by

\begin{equation}\label{8d bimodal}
\begin{split}
	\bm{\mu}(\x) = \bm{D}(\x)\nabla \log\big(\beta_7\mathcal{N}(\x; \bm{\mu}_7, \bm{\Sigma}_7) + \beta_{8}\mathcal{N}(\x; \bm{\mu}_{8}, \bm{\Sigma}_{8})\big),\quad
	\bm{D}(\x) = \bm{I}_8.
\end{split}
\end{equation}
We use the mixture weights of the second two-dimensional example and construct block-diagonal covariance matrices from $\bm{\Sigma}_{3}$ and $\bm{\Sigma}_{4}$ in Eq.~\eqref{2d bimodal II parameter}. The component means and block scalings are specified below.
\begin{equation}
\begin{split}
    \beta_7 = 0.55, \quad& \bm{\mu}_7 = (-1, -1, -0.3, -0.3, -0.4, -0.4, -1.6, -1.6),\\
    \beta_{8} = 0.45, \quad& \bm{\mu}_{8} = (2, 2, 0.6, 0.6, 0.8, 0.8, 2.3, 2.3),\\
    \bm{\Sigma}_7 &= \text{diag}\big\{\bm{\Sigma}_3, 0.6\bm{\Sigma}_3, 0.8\bm{\Sigma}_3, 1.2\bm{\Sigma}_3\big\},\\
    \bm{\Sigma}_{8} &= \text{diag}\big\{0.5\bm{\Sigma}_4, 0.3\bm{\Sigma}_4, 0.4\bm{\Sigma}_4, 0.6\bm{\Sigma}_4\big\}.
\end{split}
\end{equation}
The exact solution is $p^*(\x) = \beta_7\mathcal{N}(\x; \bm{\mu}_7, \bm{\Sigma}_7) + \beta_{8}\mathcal{N}(\x; \bm{\mu}_{8}, \bm{\Sigma}_{8})$.
	
	We set $K=8$ and use $L=10$ affine coupling layers per outer stage, each parameterized by a network with two hidden layers of \(80\) neurons each. We start with $N = 50{,}000$ training points with a batch size of $m_\text{b} = 1{,}000$. Training comprises $N_\text{a} = 20$ adaptive iterations and $N_\text{s} = 50$ epochs per iteration. The learning rate is fixed at $\eta = 0.0001$ throughout training. For SRAS, we set the mixture weights to $\gamma_0 = 0.05$, $\gamma_1 = 0.1$, $\gamma_2  = 0.6$, and $\gamma_3 = 0.25$. The quantile parameter is $\delta = 0.05$, and the box expansion factor is $\rho = 0.05$. We generate $N_\text{c} = 50{,}000$ candidate points and select $m = 2{,}000$ with the largest residual magnitudes as Gaussian mixture centers. The Gaussian standard deviation is $\sigma = 0.1$.
	
\begin{figure}[h]
	\centering
	\includegraphics[width=0.70\textwidth]{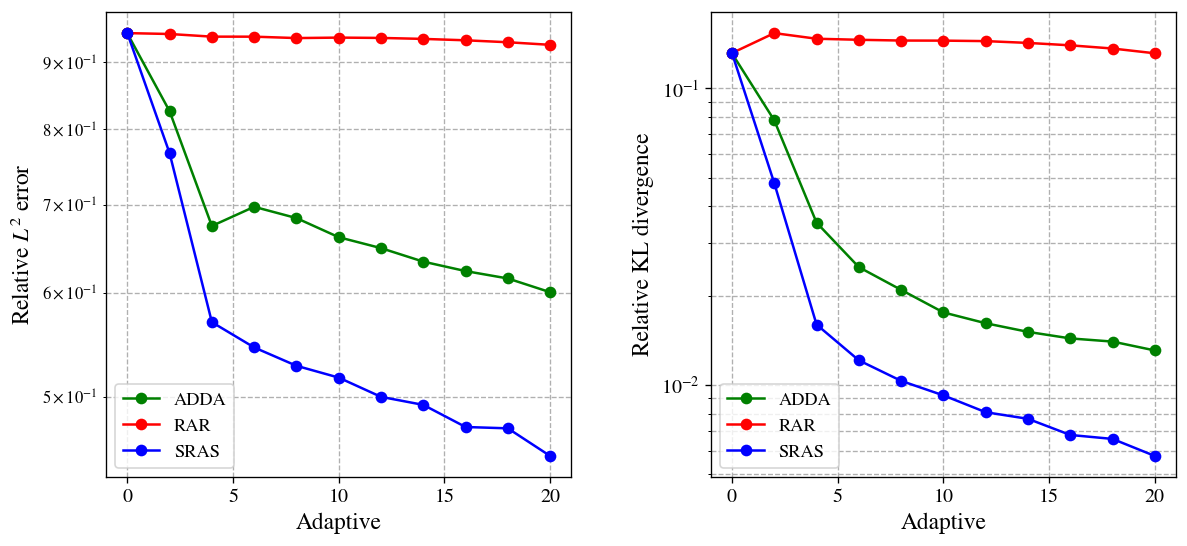}
	\caption{Evolution of the relative $L^2$ error and relative KL divergence for different sampling strategies.}
	\label{8d bimodal errors}
\end{figure}

	Figure \ref{8d bimodal errors} shows the evolution of the relative $L^2$ error and relative KL divergence over the adaptive iterations. ADDA reduces both metrics overall, despite an early fluctuation in its relative \(L^2\) error. RAR achieves only a small reduction in the relative \(L^2\) error, while its relative KL divergence remains near its initial level. SRAS achieves the largest overall reductions and attains the lowest final errors among the three methods. Nevertheless, its relative $L^2$ error remains substantial, even though its relative KL divergence falls below $10^{-2}$. These metrics emphasize different aspects of the approximation. In particular, the Monte Carlo estimate of the $L^2$ error involves importance weights that can make it sensitive to samples in low-density regions, whereas the KL divergence measures discrepancies through the log-density ratio averaged under the reference distribution. For high-dimensional problems, these metrics may therefore be supplemented by application-specific quantities, such as moments or marginal distributions.

\section{Conclusions}\label{Conclusions}
	We have proposed Solution-Residual Adaptive Sampling (SRAS) for solving stationary Fokker-Planck equations on unbounded domains using KRnet. The normalizing-flow representation ensures normalization and non-negativity by construction, while SRAS combines solution-induced sampling with residual-guided Gaussian mixture sampling to allocate collocation points. The current density approximation also defines the candidate region for residual exploration. Numerical experiments on five problems in two to eight dimensions show that SRAS generally reduces the relative $L^2$ error and relative KL divergence in fewer adaptive iterations than the ADDA and RAR implementations considered. It achieves the lowest reported errors in both two-dimensional examples and the lowest final errors in the three higher-dimensional examples. These results support combining solution-induced sampling with residual-guided refinement to improve density approximation under the tested settings. The experiments also indicate directions for further work. The preferred Gaussian exploration scale varies between problems, motivating an adaptive scale-selection strategy. SRAS improves on the baselines in the eight-dimensional example, while further improvement in relative $L^2$ accuracy remains possible. Future work will explore adaptive sampling strategies for higher-dimensional SFPEs and assess solution quality using application-specific quantities, such as moments and marginal distributions.

\bibliographystyle{plain}
\bibliography{references}

\appendix
\section{KRnet}\label{Appendix A}
	This appendix summarizes the KRnet architecture adopted in this work \cite{KRnet}. We omit the nonlinear layer to satisfy the smoothness requirements discussed in Section \ref{sec:NF_framework}. The resulting transformation consists of $K-1$ outer stages, each containing a sequence of inner transformations.
\begin{align}
	\f_{\mathrm{KR}}(\z) &= \big(\f_{[K-1]}^{\mathrm{out}} \circ \cdots \circ \f_{[1]}^{\mathrm{out}}\big)(\z), \\
	\f_{[k]}^{\mathrm{out}}(\z) &= \big(L_{\text{S}, k} \circ \f_{[k, L]}^{\mathrm{in}} \circ \cdots \circ \f_{[k, 1]}^{\mathrm{in}} \circ L_{\text{R}, k}\big)(\z),
\end{align}
where $\f_{[k]}^{\mathrm{out}}$ denotes the transformation at the $k$-th outer-loop stage, and each $\f_{[k,l]}^{\mathrm{in}}$ combines an affine coupling layer with a scale and bias layer. The operators $L_{\text{S}, k}$ and $L_{\text{R}, k}$ denote the squeezing and rotation layers, respectively.

Let $\z_{[0]}=\z$ and define
\begin{equation}
	\z_{[k]} = (\f_{[k]}^{\mathrm{out}}\circ\cdots\circ\f_{[1]}^{\mathrm{out}})(\z), \quad k=1,\dots,K-1 .
\end{equation}
For simplicity, assume an equal partition with $d=Kq$ across stages:
\[
\z_{[k]}=
\bigl(\z_{[k]}^{(1)},\ldots,\z_{[k]}^{(K)}\bigr),
\qquad
\z_{[k]}^{(i)}\in\mathbb R^q.
\]
The partition structure remains fixed, while successive outer stages deactivate one partition at a time. At stage \(k\), the rotation, scale and bias, and affine coupling layers act only on the first \((K-k+1)\) partitions. Previously deactivated coordinates are carried forward unchanged.
In the layer descriptions below, $\hat{\z}\in\mathbb R^{d_{\text{a} }}$
denotes the active input and $\hat{\bm y}\in\mathbb R^{d_{\text{a}}}$
the corresponding output, where $d_{\text{a}}$ is the number
of active coordinates.

	\tbf{Rotation layer.} The rotation layer applies a trainable linear transformation:
\begin{equation}
	\hat{\bm y}=W\hat{\z}.
\end{equation}
The matrix \(\bm{W}\) is parameterized through an LU factorization, with the diagonal entries of the triangular factors constrained to be nonzero to preserve invertibility.
	
	\tbf{Scale and bias layer.} This layer applies an element-wise affine transformation:
\begin{equation}
	\hat{\bm y}=\bm a\odot\hat{\z}+\bm b,
\end{equation}
where \(\bm a,\bm b\in\mathbb R^{d_{\mathrm a}}\) are trainable scaling and bias vectors, respectively, and $\odot$ denotes the Hadamard product.
Every component of $\bm a$ is constrained to be nonzero.

	\tbf{Affine coupling layer.} Partition the active input as \(\hat{\z}=(\hat{\z}_1,\hat{\z}_2)\), where \(\hat{\z}_1\in\mathbb R^s\) and \(\hat{\z}_2\in\mathbb R^{d_{\mathrm a}-s}\). The layer applies an invertible affine transformation on one partition conditioned on the other:
\begin{equation}\label{ACL}
	\begin{aligned}
    \hat{\bm y}_1 &= \hat{\z}_1,\\
    \hat{\bm y}_2 &=\hat{\z}_2\odot\left[\bm{1}_{d_{\mathrm a}-s}+\alpha\tanh\bigl(\bm{S}(\hat{\z}_1)\bigr)\right]+\exp(\bm\beta)\odot\tanh\bigl(\bm{T}(\hat{\z}_1)\bigr).
	\end{aligned}
\end{equation}
where \(\bm{S},\bm{T}: \mathbb R^s\to\mathbb R^{d_{\mathrm a}-s}\) are parameterized by a neural network through \(\bigl(S(\hat{\z}_1),T(\hat{\z}_1)\bigr)=\mathrm{NN}(\hat{\z}_1)\).
Here, $\alpha\in(0, 1)$ is a hyperparameter, $\bm{\beta} \in\R^{d_{\text{a}}- s}$ is trainable, and the exponential and hyperbolic tangent functions are applied componentwise.
	
	The affine coupling layer updates only one partition while keeping the other fixed. By alternating the two partitions across successive coupling layers, all components can be transformed.

	\tbf{Squeezing layer.}  At the end of outer stage \(k\), the \((K-k+1)\)-th partition is deactivated. The remaining active coordinates are specified by the mask 
\begin{equation}
	\bm{q}_{[k]} = (\underbrace{1, \dots, 1}_{n_{[k]}}, \underbrace{0, \dots, 0}_{d - n_{[k]}}), \qquad n_{[k]}=(K-k)q.
\end{equation}
The coordinates selected by \(\bm q_{[k]}\odot\hat{\z}_{[k]}\) remain active, whereas those selected by \((\bm 1_d-\bm q_{[k]})\odot\hat{\z}_{[k]}\) are unchanged in subsequent outer stages.

\end{document}